\documentclass{article}
\usepackage{spconf,amsmath,amssymb,amsfonts,graphicx}

\usepackage{cite}
\usepackage{textcomp}
\usepackage{xcolor}
\usepackage{booktabs,tabularx,array}
\usepackage{enumitem}
\setlist{nosep,leftmargin=14pt}
\definecolor{paperblue}{rgb}{0.21,0.49,0.74}
\usepackage{xurl}
\usepackage{microtype}

\newcommand{\projectrepo}{\url{https://github.com/kaanbuyukdemirci/isbi-model-resolution-comparison}}
\newcommand{\PlannedModelResolutionPairs}{108}
\newcommand{\CurrentRetainedRuns}{57}
\newcommand{\CurrentExcludedRuns}{51}
\newcommand{\CurrentBestMap}{0.709}

\newcommand{\CurrentBestResolution}{1600\!\times\!1280}
\newcommand{\CurrentBestModel}{DINO/ConvNeXt-S}
\newcommand{\CurrentLowestMedianTrainHours}{0.45}
\newcommand{\CurrentHighestMedianTrainHours}{1.40}
\newcommand{\CurrentLowestMedianInferenceMs}{23.6}
\newcommand{\CurrentHighestMedianInferenceMs}{113.7}
\newcommand{\CurrentSmallBoxLowestMap}{0.367}
\newcommand{\CurrentSmallBoxHighestMap}{0.441}

\newcommand{\CurrentSmallBoxDelta}{0.074}
\newcommand{\CurrentMiddleBoxDelta}{0.007}
\newcommand{\CurrentLargeBoxDelta}{0.032}
\newcommand{\CurrentLowestMedianMap}{0.626}

\newcommand{\CurrentHighestMedianMap}{0.660}

\newcommand{\CurrentLowestResHighestMap}{0.670}
\newcommand{\CurrentHighestResHighestMap}{0.709}

\title{Impact of Input Resolution on Mammographic Mass Detection:\\A Controlled Multi-Architecture Study}

\name{Kaan Buyukdemirci$^{\star\dagger}$ \qquad Emine Ulku Saritas$^{\star\dagger}$}
\address{$^{\star}$Department of Electrical and Electronics Engineering, Bilkent University, Ankara 06800, Turkey\\
$^{\dagger}$National Magnetic Resonance Research Center (UMRAM), Bilkent University, Ankara 06800, Turkey\\
Email: k.buyukdemirci@bilkent.edu.tr, saritas@ee.bilkent.edu.tr}

\begin{document}
\ninept

\maketitle

\begin{abstract}
Deep-learning-based mass detectors usually downsample mammograms, trading fine image detail for lower computational cost. However, published studies cannot reveal how much accuracy this trade-off sacrifices, since they also differ in architecture, preprocessing, and evaluation protocol.
In this work, we isolated the effect of input resolution in a controlled study on the VinDr-Mammo dataset, training 18 detectors at six input resolutions ranging from $512\!\times\!416$ to $\CurrentBestResolution$ pixels, for \PlannedModelResolutionPairs{} model–resolution combinations. Whole mammograms were resized without cropping or tiling. For each model, preprocessing, model initialization, augmentation, and evaluation procedures were kept fixed across resolutions. Detection accuracy was evaluated by mAP$_{50}$ on the 219 mass-positive images of the official test split, and \CurrentRetainedRuns{} combinations with test mAP$_{50}\geq0.60$ were analyzed. From the lowest to the highest resolution, median mAP$_{50}$ rose from \CurrentLowestMedianMap{} to \CurrentHighestMedianMap{} and the best mAP$_{50}$ from \CurrentLowestResHighestMap{} to \CurrentHighestResHighestMap{}. The best result, achieved by DINO with a ConvNeXt-S backbone, is the highest mAP$_{50}$ reported to date on mass-positive VinDr-Mammo images. Gains were concentrated in small masses: for the smallest third of lesions, median mAP$_{50}$ rose by \CurrentSmallBoxDelta{} (from \CurrentSmallBoxLowestMap{} to \CurrentSmallBoxHighestMap{}), compared with \CurrentMiddleBoxDelta{} and \CurrentLargeBoxDelta{} for medium and large masses. Median training time increased from \CurrentLowestMedianTrainHours{} to \CurrentHighestMedianTrainHours{} h, and median inference time from \CurrentLowestMedianInferenceMs{} to \CurrentHighestMedianInferenceMs{} ms per image. Therefore, higher input resolution mainly benefits the small masses that are hardest to detect, at a computational cost that remains practical.

\end{abstract}


\section{Introduction}
\label{sec:intro}

Deep-learning methods have been used to support mammogram interpretation \cite{shen2019screening,wu2020radiologists}. Applications include image-level classification to identify mammograms with suspicious findings and mass detection to localize individual lesions. Mammograms can contain a large number of pixels, making full-resolution processing computationally demanding. As an example, the publicly available VinDr-Mammo dataset features full mammograms that can be as large as $3580\times2812$ pixels, along with annotated breast findings \cite{nguyen2023vindr}. To reduce memory use and processing time, mammograms are often resized to a lower input resolution. This may remove details needed to detect small or hard-to-see masses. Using a higher input resolution can preserve more detail, but increases the computational load. The question is therefore how much higher resolution improves detection and whether the improvement justifies the cost.

Previous studies have investigated
mass detection by dividing mammograms
into small high-resolution patches to preserve
fine details while reducing the computational load
\cite{rangarajan2022ultrahigh,bulatovic2025patched}.
Other studies have examined resizing, cropping, and
preprocessing for mammogram classification
\cite{petrini2026resolution,ferreira2026preprocessing,
huynh2023roi}. Overall, the studies in the literature differ in their models, training
procedures, image preparation, and test images
\cite{trang2025manga,ghosh2024mammoclip}.
As a result, an important question that remains is how the input resolution alone affects mass
detection performance.

In this work, we compared 18 models at six different input resolutions using the 219 mass-positive images of the official VinDr-Mammo test split. For
each model,
the preprocessing, model initialization,
augmentation, and evaluation procedures were kept fixed, while only the input resolution was changed. At each resolution, full
mammograms were resized without cropping or dividing into
patches. The detection performance was evaluated using mean
average precision at a bounding-box overlap threshold
of 0.50 ($\mathrm{mAP}_{50}$). The performances at different mass sizes were also compared, along with training and inference times.
Our contributions are:
\begin{enumerate}
\item  A controlled comparison of 18 detection models
at six different input resolutions, covering 108
model--resolution combinations, showing that the highest mAP$_{50}$ among all the models increased from \CurrentLowestResHighestMap{} to
\CurrentHighestResHighestMap{} between the lowest input resolution at $512\!\times\!416$ and the highest input resolution at $1600\!\times\!1280$. 
\item An mAP$_{50}$ score of \CurrentBestMap{} achieved using
\CurrentBestModel{} at $\CurrentBestResolution$ resolution, the
highest reported score in the literature 
using mass-positive
VinDr-Mammo images (see Table~\ref{tab:prior_comparison}). 
\item An analysis of mAP$_{50}$ score by mass size, showing that the largest performance gains were achieved for the smallest masses.
\end{enumerate}
The code,
configurations, and other resources needed to
reproduce the experiments will be released at \projectrepo.

\section{Data and Methods}
\label{sec:dataset}

\subsection{Dataset and Preprocessing}

\begin{figure*}[t]
\centering
\includegraphics[width=0.85\textwidth]{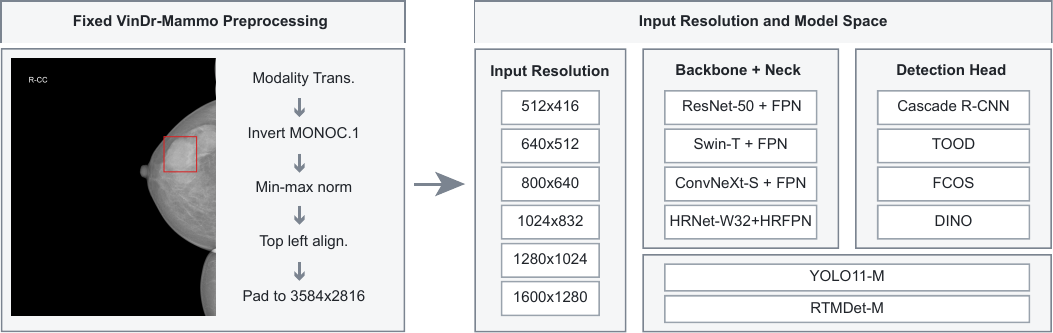}
\caption{Following preprocessing of VinDr-Mammo images, mass detection performances of 18 models are evaluated at six input resolutions. The model--resolution combinations are as listed. Four backbones are combined with four detection networks, resulting in 16 models. DINO uses its
own channel mapper instead of FPN or HRFPN.
The other 2 models, YOLO11-M and RTMDet-M, fully retain their native architectures.}
\label{fig:study_design}
\end{figure*}

VinDr-Mammo v1.0.0 contains 20,000 mammograms from 5,000 exams, with a total of 1,226 mass boxes annotated on 1,113 images \cite{pham2022vindrphysionet}.
Here, we retained the official 1,000-exam test split and held out 15\% of training
exams for validation. The resulting training, validation, and test sets contained 743, 151, and 219 mass-positive images, respectively. Then, 186 mass-negative images were added to the training set to help the models distinguish masses from normal breast tissue. Both the MLO and CC views of a breast had to lack a mass annotation
for either to qualify as negative. 

First, the DICOM images were converted to floating-point arrays, with the image polarity corrected when necessary and pixel values normalized to $[0,1]$. No cropping, mirroring, histogram equalization, CLAHE, 
lesion cropping, or tiling was applied. The mammograms were zero padded to $3584\!\times\!2816$ pixels, the nearest dimension divisible by 32. 
These images were then resized while preserving the aspect ratio. Compatibility with different network architectures was ensured by applying minimal right padding to the nearest multiple of 32. Six different input resolutions were considered: 
$512\!\times\!416$, $640\!\times\!512$, $800\!\times\!640$,
$1024\!\times\!832$, $1280\!\times\!1024$, and $1600\!\times\!1280$. 
Bounding-box coordinates were scaled by the same factor at each
resolution. Grayscale values were converted to RGB values by copying pixel intensities into three channels 
to match the input format expected by the models 
(Fig.~\ref{fig:study_design}).

\begin{figure*}[t]
\centering
\begin{minipage}[t]{\columnwidth}
    \centering
    \includegraphics[width=0.96\linewidth]{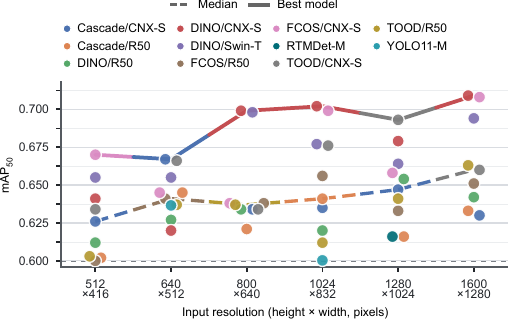}\par
    \textbf{(a)} Resolution vs. mAP$_{50}$
\end{minipage}\hfill
\begin{minipage}[t]{\columnwidth}
    \centering
    \includegraphics[width=0.96\linewidth]{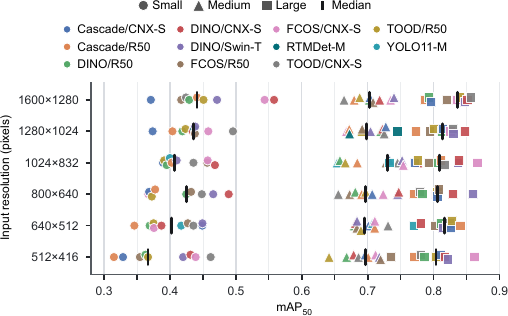}\par
    \textbf{(b)} mAP$_{50}$ by mass size and resolution
\end{minipage}
\caption{Comparison of detection performances at different input resolutions. 
(a) mAP$_{50}$ scores of models at different resolutions. Dashed and solid lines connect the median and
best mAP$_{50}$ scores per resolution, respectively. (b) mAP$_{50}$ scores for three different mass-size groups at different resolutions. Markers indicate mass sizes: circle for small,
triangle for medium, and square for large masses, with
vertical ticks marking group medians. Colors identify models in both (a) and (b). Only the results for model--resolution combinations with mAP$_{50}\geq0.60$ are shown.}
\label{fig:map_results}
\end{figure*}

\subsection{Controlled Resolution Sweep}
\label{sec:methods}

We evaluated four feature extraction networks (i.e., backbones):
ResNet-50, Swin-T, HRNet-W32, and ConvNeXt-S. 
We combined each backbone with four detection networks:
Cascade R-CNN, TOOD, FCOS, and DINO
\cite{cai2018cascadercnn,feng2021tood,tian2019fcos,
zhang2023dino}, resulting in 16 detection models. 
Independent of the choice of backbone, DINO used its own channel mapper. In all other cases, ResNet-50, Swin-T, and ConvNeXt-S were paired with an FPN neck, while HRNet-W32 was paired with an HRFPN neck.
In addition, we evaluated YOLO11-M and RTMDet-M, each using its native backbone, neck, and
detection head. We used implementations
from MMDetection and Ultralytics
\cite{chen2019mmdetection,ultralytics2024yolo11}.
All 18 models processed full mammograms, without tiling,
cross-view fusion, test-time augmentation, or ensembling.

Increasing input resolution 
$H\times W$ increased feature-map locations but not learned parameter shapes.
At stride $s$, the feature map contained approximately
$\lceil H/s\rceil\!\times\!\lceil W/s\rceil$ locations. Hence, resolution changed
spatial sampling while channel widths and pyramid levels remained fixed.
Parameter counts ranged from 20.1M to 94.5M across the models and remained fixed within each model.

In the end, the controlled multi-architecture study contained 18 models and six resolution levels, for a total of $\PlannedModelResolutionPairs{}$
model--resolution combinations. Each model--resolution combination used five learning-rate and
batch-size settings, giving 540 planned runs. Augmentation settings,
initial weights, random seed, and optimizer 
remained
the same. 
Baseline learning rates and two batch sizes were selected for each
model at $640\!\times\!512$ using validation only. With the smaller
batch size, we tested learning rates of 0.5, 1, and 1.5 times the baseline rate.
With the larger batch size, we tested 0.5 and 1 times the baseline rate.
We initialized each model using official ImageNet or COCO weights, with the one-class output layer reinitialized and input
normalization unchanged across resolutions. We used gradient accumulation to preserve the effective batch size when GPU memory required a smaller physical batch.

All runs used seed 42, one RTX 3090, at most 25 epochs, moderate geometric and
intensity augmentation ($\pm2^\circ$ rotation, 2\% translation, scaling from
0.95 to 1.05, mild intensity changes, and low-variance Gaussian noise), and early
stopping after ten epochs without a 0.002 validation-mAP$_{50}$ improvement. For each model–resolution combination, we selected the checkpoint with the highest validation mAP$_{50}$ across the five learning-rate and
batch-size settings. 
Only the selected checkpoint was evaluated on the test set. Both validation and test evaluation used mass-positive images, following related work \cite{ghosh2024mammoclip,bhat2026exemplar}.

\begin{table}[t]
\centering
\caption{Reported detection performances on
mass-positive VinDr-Mammo images. ``Official''
denotes the official test split. Numbers in
parentheses indicate test image count.
NR stands for ``not reported''.}
\label{tab:prior_comparison}
\small
\setlength{\tabcolsep}{3pt}
\renewcommand{\arraystretch}{1.12}
\begin{tabularx}{\columnwidth}{
@{}>{\raggedright\arraybackslash}p{0.48\columnwidth}
c
>{\raggedright\arraybackslash}X@{}
}
\toprule
Study & mAP$_{50}$ & Test set \\
\midrule
\textbf{This work} & \textbf{\CurrentBestMap} &
Official (219) \\

Bhat et al. \cite{bhat2026exemplar} & 0.700 &
Official (NR) \\

Trang et al. \cite{trang2025manga} & 0.690 &
Random (NR) \\

Bulatovi\'c et al. \cite{bulatovic2025patched} & 0.671 &
Official (NR) \\

Mammo-CLIP/FM
\cite{ghosh2024mammoclip,ghosh2025mammofm} & 0.580 &
Official (219) \\
\bottomrule
\end{tabularx}
\end{table}

\subsection{Quantitative Evaluations}


For quantifying detection accuracy, mAP$_{50}$ score was computed for each model--resolution combination on mass-positive images, excluding false positives on the 3,781 mass-negative official-test images.
We excluded \CurrentExcludedRuns{} combinations with test mAP$_{50}<0.60$,
leaving \CurrentRetainedRuns{} combinations for analysis. The number of retained models per resolution varied between 9 and 10. 

The 219 mass-positive test images contained 237 ground-truth
bounding boxes. 
In addition, we divided the bounding boxes into three approximately
equal-sized groups based on their area relative to the whole
image, using thresholds of 0.280\% and 0.709\% of the total
image area, resulting in 77 to 81 boxes per group.
We then calculated mAP$_{50}$
separately for each mass-size group to determine the performances at different mass sizes. 
Lastly, training time was measured
as wall-clock duration, while per-image inference time
was calculated as the average over the test images. 

\section{Results}
\label{sec:results}

Figure~\ref{fig:map_results}(a) shows mAP$_{50}$ scores for the models at different input resolutions. The median mAP$_{50}$ score rises from \CurrentLowestMedianMap{} at
$512\!\times\!416$ to \CurrentHighestMedianMap{} at
$1600\!\times\!1280$, increasing in a non-monotonic fashion. 
Per-resolution best mAP$_{50}$ scores are
also non-monotonic. The overall best mAP$_{50}$ score of \CurrentBestMap{} is achieved with
\CurrentBestModel{} at $\CurrentBestResolution$. Table~\ref{tab:prior_comparison} shows that this score is the highest reported detection accuracy on mass-positive VinDr-Mammo images in the literature, exceeding the next-highest reported score of 0.700.

As shown in Fig.~\ref{fig:map_results}(b), the models have considerably lower median mAP$_{50}$ for small masses compared with medium and large masses. Nonetheless, the small masses benefit the most from increased resolution: the median mAP$_{50}$ for the small mass sizes increases by \CurrentSmallBoxDelta{} (from \CurrentSmallBoxLowestMap{} to
\CurrentSmallBoxHighestMap{}), when compared to 
\CurrentMiddleBoxDelta{} and \CurrentLargeBoxDelta{} for the medium and large mass sizes, respectively.

\begin{figure}[t]
    \centering
    \includegraphics[width=0.96\columnwidth]{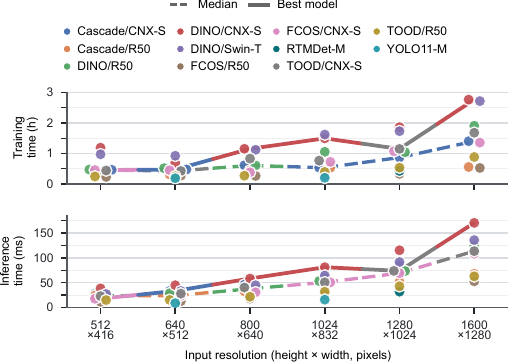}
    \caption{Training and inference times at different input resolutions. 
    Dashed lines track the median times and solid lines track the times for
    best-mAP$_{50}$ models from Fig.~\ref{fig:map_results}(a). Colors identify models.}
    \label{fig:compute_results}
\end{figure}

The median training time increases from \CurrentLowestMedianTrainHours{} to
\CurrentHighestMedianTrainHours{} h and inference time increases from
\CurrentLowestMedianInferenceMs{} to \CurrentHighestMedianInferenceMs{} ms
(Fig.~\ref{fig:compute_results}), demonstrating
a trade-off between detection accuracy and computational cost, as expected. 

\begin{figure*}[t]
    \centering
    \includegraphics[width=\textwidth]{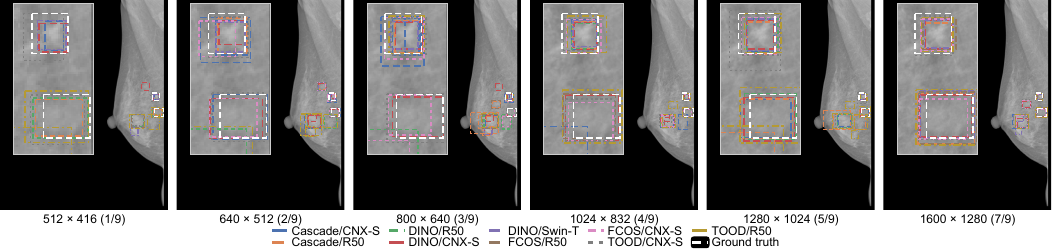}
    \caption{Representative detection results across all six resolutions on a mammogram with two adjacent small masses. 
    Model predictions and ground truth boxes are shown, with rectangular insets displaying close-up views of the two masses. The number of models successfully detecting both masses increases with increasing input resolution, from 1/9 at $512\times 416$ to 7/9 at $1600\times 1280$ (confidence $\geq 0.15$, $\text{IoU} \geq 0.50$).}
    \label{fig:resolution_comparison}
\end{figure*}

Figure~\ref{fig:resolution_comparison} shows an example case with two adjacent
small masses (sizes of 0.20\% and 0.15\% of the total image area). The number of models successfully detecting both
masses increases at every resolution step, from 1/9 at $512\!\times\!416$ to 7/9 at
$1600\!\times\!1280$. At a fixed feature-map stride, a higher-resolution input gives
each mass more pixels and feature-map samples without increasing the parameter count. Consistent with the performance gain observed for smaller masses, this example
illustrates how retaining high-resolution image details can help detect and distinguish nearby masses.  

\section{Discussion and Conclusion}
\label{sec:conclusion}

This work presented a controlled comparison of detection performances across models by varying input resolution within fixed detector and training setups. 
The largest
tested input resolution produces both the highest median mAP$_{50}$ and the best
individual result. 
To our knowledge, the best mAP$_{50}$ of \CurrentBestMap{}, obtained with DINO/ConvNeXt-S at $1600\!\times\!1280$, is the highest reported on mass-positive VinDr-Mammo images (Table~\ref{tab:prior_comparison}). This result was achieved with a standard whole-image detector, without architectural modifications, patch-based inference, or external mammography-specific pretraining, yet it exceeds methods that rely on these strategies.

The mass-size dependent performance analysis helps explain the benefit of using higher input resolution.
From the lowest to the highest input size, the smallest masses gain
\CurrentSmallBoxDelta{} in median mAP$_{50}$, which is considerably higher than that for the larger
masses. Together with the adjacent-mass example in Fig.~\ref{fig:resolution_comparison}, this finding supports
retaining lesion detail as a reason for the improvement. Small masses still
have the lowest median scores at every input size, so increasing resolution
does not remove the mass-size dependence of performance. 


While the computational cost increases with resolution, the one-time training time remains manageable even at the highest tested resolution. In addition, the inference time remains short enough to not impose any bottlenecks in a radiological evaluation pipeline. Considering the importance of early detection and that the largest gains are observed for the smallest masses, the increase in computational cost might be well justified. 



\section{Compliance with ethical standards}
\label{sec:ethics}

This work used de-identified human
subject data from the publicly available VinDr-Mammo dataset 
\cite{pham2022vindrphysionet}.

\section{Acknowledgments}
\label{sec:acknowledgments}

The authors have no
relevant financial or non-financial interests to disclose.
Codex was used to help develop the code and extract shared repositories from larger codebases. 
All Codex-generated code was reviewed and tested in detail by the authors.

\small
\bibliographystyle{IEEEtran}
\bibliography{references}

\end{document}